# DRAGD: A Federated Unlearning Data Reconstruction Attack Based on Gradient Differences

Bocheng Ju, Junchao Fan, Jiaqi Liu, Xiaolin Chang

*Abstract*—Federated learning enables collaborative machine learning while preserving data privacy. However, the rise of federated unlearning, designed to allow clients to erase their data from the global model, introduces new privacy concerns. Specifically, the gradient exchanges during the unlearning process can leak sensitive information about deleted data. In this paper, we introduce DRAGD, a novel attack that exploits gradient discrepancies before and after unlearning to reconstruct forgotten data. We also present DRAGDP, an enhanced version of DRAGD that leverages publicly available prior data to improve reconstruction accuracy, particularly for complex datasets like facial images. Extensive experiments across multiple datasets demonstrate that DRAGD and DRAGDP significantly outperform existing methods in data reconstruction. Our work highlights a critical privacy vulnerability in federated unlearning and offers a practical solution, advancing the security of federated unlearning systems in real-world applications.

## I. INTRODUCTION

OVER recent years, federated learning (FL) [1]-[6] has emerged as a transformative paradigm for collaborative machine learning, enabling multiple clients to jointly train a global model without sharing their raw data. This decentralized approach is increasingly favored for its ability to enhance data privacy and security in sensitive domains such as healthcare, finance, and mobile applications. However, clients participating in FL may, for various reasons—including privacy concerns, regulatory requirements, and data fidelity—request the complete erasure of their data from trained models. Landmark privacy legislations, such as the General Data Protection Regulation (GDPR) [6] and the California Consumer Privacy Act (CCPA) [8], have institutionalized the "right to be forgotten," granting users the legal entitlement to have their personal data and its influence erased from deployed machine learning systems upon request.

In response, the concept of federated unlearning (FU) has been proposed, which aims to empower the FL server to effectively eliminate the influence of specific data points from the global model when a client requests data removal [9]. Despite the rapid progress in efficient unlearning algorithms, federated unlearning introduces a series of new challenges [10]. Notably, it remains unclear whether the data removal process itself could expose new privacy risks, particularly during the model update and gradient exchange phases inherent to FL protocols.

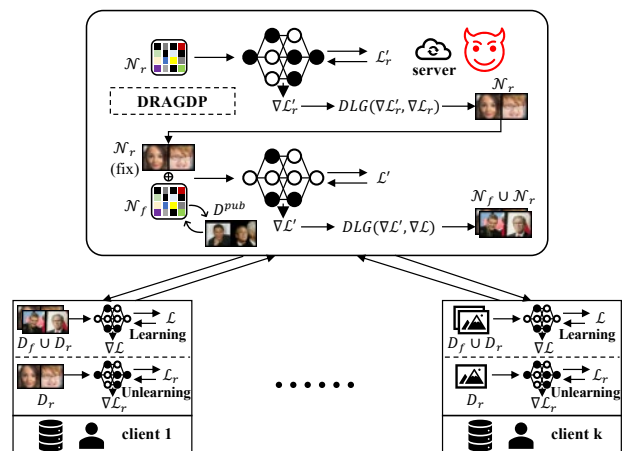

**Fig.1.** Overview of DRAGDP. The adversarial server collects client gradients before and after the unlearning process. DRAGDP reconstructs the remaining data from post-unlearning gradients and uses the difference between pre- and post-unlearning gradients to efficiently recover the target forgotten data. DRAGDP leverages public data priors to further enhance reconstruction accuracy in tasks with strong prior structures (e.g., face images).

As federated learning frameworks continue to evolve and scale, ensuring that unlearning operations comply not only with regulatory standards but also protect against potential privacy leakage and adversarial exploitation has become a critical and urgent research challenge. While existing approaches have made significant progress in improving unlearning efficiency, minimizing the residual impact of deleted data, and maintaining model performance on the remaining dataset, most mainstream methods focus on retraining the model after data removal or accelerating the retraining process to quickly "forget" the targeted information [11]-[14]. While these methods have achieved significant progress in unlearning speed and utility preservation, they overlook a fundamental privacy risk inherent to the unlearning process itself.

The core problem lies in gradient discrepancies. During federated unlearning, the gradient exchanges before and after data removal create distinctive patterns that can reveal sensitive information about the supposedly "forgotten" data. Unlike

All the authors are with School of Computer and Information Technology, Beijing Jiaotong University, Beijing, 100044, China. E-mails: {24115070, xlchang}@bjtu.edu.cn.

standard federated learning where gradients may leak some information, federated unlearning amplifies this risk by providing attackers with comparative gradient information [15]-[17]. These gradient discrepancies act as a privacy fingerprint of the removed data, enabling sophisticated reconstruction attacks.

Current privacy-preserving defenses—including differential privacy and secure aggregation—are insufficient for this unique threat. They either impose substantial computational overhead, degrade model performance, or fail to address the specific vulnerabilities introduced by gradient discrepancies in federated unlearning [18], [19]. This leaves federated unlearning systems exposed to attacks that can recover the very data intended to be forgotten.

This gap between privacy expectations and actual security represents a critical vulnerability in federated unlearning systems. As these systems scale and handle increasingly sensitive data, understanding and mitigating privacy risks from gradient discrepancies becomes essential for practical deployment and regulatory compliance.

In this paper, we propose a new framework for both attacking and defending privacy in federated unlearning by introducing a set of gradient difference-based methods. We first design **DRAGD**, i.e., <u>D</u>ata <u>R</u>econstruction <u>A</u>ttack based on <u>G</u>radient <u>D</u>ifference, a novel attack that, for the first time, systematically exploits the gradient discrepancies before and after the unlearning operation to reconstruct the actual forgotten data in federated learning systems. Unlike previous work, which mainly addresses membership inference or general privacy leakage, DRAGD directly targets the challenge of data reconstruction, uncovering the true extent of privacy risks exposed during the unlearning process.

Building on this, we further propose **DRAGDP**, i.e., <u>D</u>ata <u>R</u>econstruction <u>A</u>ttack based on <u>G</u>radient <u>D</u>ifference <u>P</u>ro, which leverages publicly available prior data to initialize and guide the reconstruction process, significantly boosting the attack's effectiveness for data with strong prior structures—such as face images—where traditional gradient attacks would struggle. This dual-stage approach allows us to reveal and quantify a more severe privacy risk than previously reported in federated unlearning literature.

The main contributions of this paper are summarized as follows:

- **Novel Attack Framework:** We propose DRAGD and DRAGDP, two novel data reconstruction attack methods for federated unlearning, based on the analysis of gradient differences. By systematically exploiting the gradient updates before and after the unlearning process, DRAGD effectively reconstructs forgotten data, while DRAGDP further enhances attack performance by incorporating prior knowledge from public datasets, especially for complex data types such as facial images.
- **Comprehensive Experimental Validation:** We conduct extensive experiments on multiple benchmark datasets and various model architectures to validate our proposed methods. Comparative experiments demonstrate that DRAGD and DRAGDP substantially outperform existing attacks in data reconstruction quality. Additionally, ablation studies validate the necessity of each component in our framework, confirming the correctness and effectiveness of our approach.

This paper has the following structure. Section II reviews related work. Section III defines the threat model. Section IV describes our attack methodology, and Section V details the experimental evaluation. Finally, Section VI concludes the paper.

## II. RELATED WORKS

### A. Federated Unlearning

Federated unlearning is an emerging paradigm that effectively updates a FL model after completely removing data without requiring it to be retrained from scratch. Existing federated unlearning studies [12], [13] mainly concentrate on designing effective methods to erase a client's data. For instance, Liu et al. [9] take the first step by proposing FedEraser, the first federated unlearning methodology that can eliminate the influence of a client's data on the global FL model while significantly reducing the time used for constructing the unlearned FL model. Wu et al. [20] further eliminate a client's contribution by subtracting the accumulated historical updates from the model and leveraging the knowledge distillation technique to restore the model's performance without relying on any data from the clients. In addition, Wang et al. [21] observe that different channels have a varying contribution to different categories in image classification tasks and then propose a method for removing the information about particular categories. Liu et al. [22] propose a smart retraining method for federated unlearning without communication protocols.

### B. Data Reconstruction Attack

Data reconstruction attack [23], [24], [25] represent a prominent class of privacy attacks in FL, where adversaries leverage shared gradients to recover data labels and reconstruct the corresponding input samples. Zhu et al. [23] first proposed this attack paradigm by formulating it as an optimization problem aimed at minimizing the gradient discrepancy between actual and synthetic images to recover the original data. Building on this, Zhao et al. [24] introduced iDLG, a straightforward single-label recovery approach that significantly improved reconstruction performance.

Geiping et al. [25] introduced the use of the Adam optimizer in this context and replaced the traditional L2 loss with a cosine similarity-based objective, enabling high-fidelity image recovery from high-resolution datasets. Yin et al. [26] extended the attack by incorporating additional regularization and a batch-based label recovery algorithm, which allowed for the reconstruction of multiple input samples in a batch.

To address the challenges of gradient inversion attacks under larger batch sizes and higher-resolution settings, researchers have increasingly incorporated generative models as powerful priors. GAN-based approaches have led this direction, with Jeon et al. [27] pioneering the use of GANs to improve reconstruction quality, followed by Li et al. [28] who combined GAN priors with gradient-free optimizers to bypass existing defenses. Fang et al. [29] further enhanced this approach by improving GAN latent space representations for better attack performance.Beyond traditional GANs, Yang et al. [30] proposed a generative gradient inversion framework that eliminates the need for iterative optimization through auxiliary data and feature separation techniques. Most recently, Wu et al. [31] introduced DGGI, which leverages diffusion models as stronger image priors and incorporates consistency regularization to address spatial misalignment issues in reconstructions.

Unlike previous methods that rely on gradient inversion and generative models for data reconstruction, our work leverages gradient discrepancies before and after unlearning, achieving more precise and efficient data reconstruction.

*C. Privacy Preservation in FL*

Existing research efforts for achieving privacy preservation in FL can be generally categorized into cryptographybased and gradient-degradation-based approaches.

A common type of cryptographic solution is secure multi-party computation (MPC), which aims to have a set of parties to jointly compute the output of a function over their private inputs in a way that only the intended output is revealed to the parties. This can be achieved by designing custom protocols [32], [33], or via secure aggregation schemes such as homomorphic encryption [34] and secret sharing [49], [35]. However, merely relying on MPC isn't sufficient to resist inference attacks over the output [36], [37].

Another line of research seeks to constrain the amount of leaked sensitive information by intentionally sharing degraded gradients. Differential privacy (DP) is the standard way to quantify and limit the privacy disclosure about individual users. DP can be applied at either the server's side (central DP) or the client's side (local DP). In comparison, local DP provides a better notion of privacy as it does not require the client to trust anyone. It utilizes a randomized mechanism to distort the gradients before sharing them with the server [38], [39]. DP offers a worst-case information theoretic guarantee on how much an adversary can learn from the released data. However, for these worst-case bounds to be most meaningful, they typically involve adding too much noise which often reduces the utility of the trained models. Zhang et al. [40] combined DP with gradient pruning in Federated Learning, proposing SLGP and RLGP to prevent data reconstruction from shared gradients while preserving model performance. In addition to DP, it is demonstrated that performing gradient compression/sparsification can also help to prevent information leakage from the gradients. A most recent work by Sun et al. [41] identifies the data representation leakage from gradients as the root cause of privacy leakage in FL and proposes a defense named Soteria, which computes the gradients based on perturbed data representations. It is shown that Soteria can achieve a certifiable level of robustness while maintaining good model utility.

Unlike cryptographic and gradient-degradation approaches, our work presents a lightweight defense that adaptively injects noise into sensitive gradient components, effectively preventing data reconstruction attacks while preserving model utility and avoiding the significant computational overhead common in other privacy-preserving methods.

III. THREAT MODEL

In this section, we outline the scenario we follow and clarify our underlying assumptions. Subsequently, we provide details about the attacker's goal, and capabilities.

*A. Attacker's Goal*

In our threat model, we consider a federated learning environment where multiple clients collaboratively train a shared global model under the coordination of a central server. Each client maintains its own private dataset, which remains strictly local and is never directly exposed to other participants throughout the training process. Within this context, the adversary is assumed to be the server orchestrating the federated learning procedure, operating in an "honest-but-curious" manner. That is, while the server faithfully follows the prescribed federated learning protocol and does not tamper with the standard operations of model aggregation or client updates, it seeks to extract as much information as possible from the data exchanged during training—particularly from the gradients transmitted by clients.

The primary objective of the attacker in this setting is to reconstruct sensitive user data that has been deliberately erased from the global model as part of a federated unlearning process. Federated unlearning refers to the intentional removal or unlearning of specific data points from the global model, typically motivated by privacy requirements or regulatory mandates for data deletion. During this process, gradients associated with both pre-unlearning and post-unlearning states are exchanged between clients and the server. The attacker exploits the differences in these gradients to infer the private information corresponding to the data points that were intended to be forgotten.

Importantly, the attacker does not aim to disrupt the learning process, degrade model performance, or mislead non-target clients. Instead, the attack is purely information-theoretic, focusing on extracting specific details about the forgotten data by analyzing and correlating gradient information. The ultimate goal is to reconstruct, with high fidelity, the sensitive data that users have requested to be erased, thereby posing a significant privacy risk to individuals who rely on federated unlearning for data protection.

*B. Attacker's Capability*

The attacker's capabilities are inherently limited by the federated learning framework. Although the server can access and store the gradients uploaded by the clients, it cannot directly alter the clients' local models or data. The attacker's primary advantage lies in the ability to analyze the gradient updates associated with the unlearning phase. By comparing the gradients before and after data is forgotten, the attacker can infer specific data points that have been removed from the model.

The attacker's influence is confined to the gradient information and the unlearning requests that initiate the unlearning process. This means that the attacker cannot directly manipulate the clients' training processes or interfere with their local models. However, the gradients shared by the clients provide sufficient information for the attacker to reconstruct forgotten data, thus compromising privacy. The effectiveness of the attack is dependent on the granularity and changes in the gradients, which directly relate to the forgotten data.

## IV. PRELIMINARIES

We focus on gradient-based reconstruction attacks against FL models. In this context, we consider a FL system with $K$ local clients. Each local client $i$ has a training dataset $\mathcal{D}_i$, where $i = 1,...,K$. Generally, the coordination of a central server facilitates the collaborative training of a unified model by these $K$ clients' training data. In particular, FL aims to learn a global model with its parameters $\theta$ to minimize the averaged loss as follows:

$$\min_{\theta \in \mathbb{R}^d} \frac{1}{K} \sum_{i=1}^{K} \mathcal{L}_i(\theta, \mathcal{D}_i), \quad (1)$$

where $\mathcal{L}_i(\theta, \mathcal{D}_i) = \frac{1}{|\mathcal{D}_i|} \sum_{(x,y) \in \mathcal{D}_i} \mathcal{L}(\theta, y)$ is the local training loss for client $i$, $|\mathcal{D}_i|$ is the number of training samples of client $i$, and $d$ is the dimension of the global model $\theta$.

In the federated unlearning process, FL performs the following two fundamental steps:

**Step I. Federated Learning:** During the FL training process, each client initially downloads the global model $\mathcal{M}_G(., \theta)$ from the central server. Subsequently, in each training round, every client independently trains its local model $\mathcal{M}_i(., \theta_i)$ using its own local training data $\mathcal{D}_i$. The central server then collects all the updates from the $K$ clients and proceeds to update the global model through an aggregation of these collected updates. This results in an updated model $\mathcal{M}_{i+1}(., \theta_{i+1})$ which serves as the global model for the subsequent training round. After multiple rounds of global updates, the final model parameters are obtained as $\theta^*$.

**Step II. Federated Unlearning:** Client $i$ sends data removal requests $\mathcal{D}_f$ at a certain time and the server aims to eliminate the influence of the data $\mathcal{D}_f$ from the global model $\mathcal{M}_G(., \theta^*)$. It is worth noting that in the federated learning process, the server and the clients communicate with each other using gradients $\nabla_\theta \mathcal{L}$, and client $i$ could send the updated model $\mathcal{M}_i(., \theta_i^{'})$ trained with remaining dataset $\mathcal{D}_r$ to the server. The server then collects updates from all clients as in Step I and proceeds to aggregate the updates, after multiple rounds of global updates, the final retrained global model parameters are obtained as $\theta_u$.

## V. ATTACK METHODOLOGY

In our attack, the goal of DRAGD is to exploit the gradient differences before and after the federated unlearning process to reconstruct sensitive user data that has been intentionally erased from the global model. To achieve this objective, DRAGD comprises the following two fundamental steps:

**Step I. Reconstructing the Remaining Dataset $\mathcal{D}_r$:** First, the attacker randomly initializes and feeding noise data $\mathcal{N}_r$ into the model $\mathcal{M}_G(., \theta_u)$, where $\mathcal{D}_r, \mathcal{D}_f, \mathcal{N}_r \in R^d$. Then, the attacker computes the gradient $\nabla_{\theta_u} \mathcal{L}_r'$ from the model after retraining, and minimizes the Euclidean distance between this gradient and the actual gradient $\nabla_{\theta_u} \mathcal{L}$. The parameters of noise data $\mathcal{N}_r$ are updated as follows:

$$\theta_{\mathcal{N}_r} = \theta_{\mathcal{N}_r} - \eta_r \frac{\partial \|\nabla_{\theta_u} \mathcal{L}_r' - \nabla_{\theta_u} \mathcal{L}\|^2}{\partial \theta_{\mathcal{N}_r}} \quad (2)$$

Where $\eta_r$ represents the learning rate for reconstructing the remaining dataset, and $\|\bullet\|^2$ represents the squared Euclidean distance, which streamlines computations by eliminating the need to calculate the square root of the Euclidean distance.

**Step II. Reconstructing the Unlearning Dataset $\mathcal{D}_f$:** First, the attacker randomly initialize a virtual noise dataset $\mathcal{N}_f$ with the same dimensionality as the model input, and set its batch size to be the same as the forgotten dataset $\mathcal{D}_f$. Then, combine $\mathcal{N}_f$ with the corresponding virtual data $\mathcal{N}_r$ and input them together into the pre-unlearning model $\mathcal{M}_G(., \theta^*)$ to compute the corresponding gradient $\nabla_{\theta^*} \mathcal{L}'$. The Euclidean distance between this gradient and the actual gradient $\nabla_{\theta^*} \mathcal{L}$ is used as the objective function. However, the parameters of $\mathcal{N}_r$ are fixed, and only the newly initialized virtual data $\mathcal{N}_f$ is trained. The parameters of $\mathcal{N}_f$ are updated as follows:

$$\theta_{\mathcal{N}_f} = \theta_{\mathcal{N}_f} - \eta_f (\frac{\partial \|\nabla_{\theta^*} \mathcal{L}' - \nabla_{\theta^*} \mathcal{L}\|^2}{\partial \theta_{\mathcal{N}_f \cup \mathcal{N}_r}})_{[0:|\mathcal{N}_f|]} \quad (3)$$

Where $\eta_f$ represents the learning rate for reconstructing the unlearning dataset, and $(\bullet)_{[i,j]}$ represents extracting the $i$-th to $j$-th elements from the bracketed data. To align with the code execution flow, during the attack on the gradient $\nabla_{\theta^*} \mathcal{L}$ before unlearning, both $\mathcal{N}_f$ and $\mathcal{N}_r$ are involved in the gradient calculation and placed in the optimizer. However, after the backpropagation, the gradient of the parameters $\theta_{\mathcal{N}_r}$ is set to zero, effectively keeping $\mathcal{N}_r$ fixed. This ensures that only the parameters of the newly initialized noise data $\mathcal{N}_f$ are updated. This is the complete process for the DRAGD attack algorithm. The DRAGDP attack algorithm, on the other hand, during the reconstruction of the target forgotten data $\mathcal{D}_f$, uses the locally held public dataset $\mathcal{D}_{pub}$ as the initial value to replace the randomly initialized virtual noise data $\mathcal{N}_f$. The entire DRAGD/DRAGDP process is encapsulated in **Algorithm 1**.

---
**Algorithm 1** DRAGD/DRAGDP Algorithm
---

**Input :** The pre-unlearning model parameters $\theta^*$, the retrained model parameters $\theta_u$, the pre-unlearning gradients $\nabla_{\theta^*}\mathcal{L}$, the retrained gradients $\nabla_{\theta_u}\mathcal{L}$, the learning rate for reconstructing the remaining dataset $\eta_r$, the learning rate for reconstructing the unlearning dataset $\eta_f$, the post-unlearning attack training rounds $T$, and the public dataset $\mathcal{D}_{pub}$.

**Output:** The reconstruction data $\mathcal{N}_r$ and $\mathcal{N}_f$.

1: Randomly initialize the noise data $\mathcal{N}_r$
2: **for** $i = 1, 2, ..., T$ **do**
3: $\quad \mathcal{L}_r' = \frac{1}{|\mathcal{N}_r|} \sum_{(x,y) \in \mathcal{N}_r} \mathcal{L}(\theta_u, y)$
4: $\quad \theta_{\mathcal{N}_r} = \theta_{\mathcal{N}_r} - \eta_r \frac{\partial \|\nabla_{\theta_u}\mathcal{L}_r' - \nabla_{\theta_u}\mathcal{L}\|^2}{\partial \theta_{\mathcal{N}_r}}$
5: **end for**
6: **if** DRAGDP **do**
7: $\quad \mathcal{N}_f \leftarrow \mathcal{D}_{pub}$
8: **else**
9: $\quad$ Randomly initialize the noise data $\mathcal{N}_f$
10: **end if**
11: $Concat(\mathcal{N}_r, \mathcal{N}_f)$
12: **for** $i = 1, 2, ..., T$ **do**
13: $\quad \mathcal{L}' = \frac{1}{|\mathcal{N}_f \cup \mathcal{N}_r|} \sum_{(x,y) \in \mathcal{N}_f \cup \mathcal{N}_r} \mathcal{L}(\theta^*, y)$
14: $\quad \theta_{\mathcal{N}_f} = \theta_{\mathcal{N}_f} - \eta_f (\frac{\partial \|\nabla_{\theta^*}\mathcal{L}' - \nabla_{\theta^*}\mathcal{L}\|^2}{\partial \theta_{\mathcal{N}_f \cup \mathcal{N}_r}})_{[0:|\mathcal{N}_f|]}$
15: **end for**
16: **return** $\mathcal{N}_r, \mathcal{N}_f$

---

## VI. EVALUATION

In this section, we first describe the datasets and evaluation metrics used in our experiments. We then conduct comprehensive experiments under Non-IID settings to evaluate the effectiveness of DRAGD and DRAGDP attacks. Additionally, we assess the performance of FedANI, focusing on both its defense capabilities against these attacks and its impact on the overall efficiency of federated learning. Furthermore, we perform ablation studies for DRAGD, DRAGDP, and FedANI, analyzing the contribution of each component within these attack and defense mechanisms.

### A. Experiment Setup

*1) Dataset:* We evaluate DRAGD, DRAGDP,and FedANI using three datasets in our experiments including: MNIST, CIFAR-10, and LFW.

**MNIST.** It comprises 70,000 handwritten digits presented as 28 × 28 images. It has been normalized, placing the digits at the center of the image. The dataset encompasses samples of handwritten digits spanning from 0 to 9, with each pixel represented by a binary value (0 or 1).

**CIFAR-10.** It is a widely utilized benchmark dataset for assessing image recognition algorithms. It comprises 60,000 color images sized 32 × 32, categorized into 10 classes. Notably, this dataset maintains balance, with 6,000 randomly chosen images per class. It is divided into 50,000 training images and 10,000 testing images.

**LFW.** It contains 13,000 labeled images of faces, spanning across 5,749 individuals. Each image is a color photograph, typically sized at 250 × 250 pixels, featuring faces in various poses, lighting conditions, and backgrounds. Given that high-resolution, large-sized images may result in redundant optimization parameters, leading to noise interference in the data reconstruction process, this study preprocesses the LFW dataset by uniformly resizing the input samples to 32×32 pixel.

*2) Evaluation Metrics:* We adopt the following three metrics to evaluate the performance of DRAGD, DRAGDP,and FedANI:

**Mean Squared Error (MSE).** MSE is a metric used to measure the average squared differences between corresponding pixels of two images. It quantifies the error between the original and reconstructed images, with lower MSE values indicating higher similarity and better image quality.

**Peak Signal-to-Noise Ratio (PSNR).** PSNR is an engineering term for the ratio between the maximum possible power of a signal and the power of corrupting noise that affects the fidelity of its representation. Higher PSNR values indicate better image quality.

**Structural Similarity Index (SSIM)**. SSIM quantifies the similarity between two images, considering luminance, contrast, and structure. SSIM values range from 0 to 1, with 1 indicating identical images.

*3) Baseline:* In the context of gradient-based attacks targeting data reconstruction in federated unlearning process, we employ the following three baselines:

**DLG[23].** Exploiting subtle differences in gradients before and after data removal, the DLG attack reconstructs sensitive data points by comparing gradients from the global model. Through iterative optimization of a noise vector to minimize gradient mismatches, this method effectively infers private data even after deletion, highlighting vulnerabilities in the federated unlearning process.

**IG[25].** Focusing on data points with the greatest influence on the global model, the IG attack calculates their gradient contributions. By matching gradients before and after the unlearning phase, this attack reconstructs forgotten data, demonstrating precision in identifying high-impact data points and effectively inferring sensitive information.

*CPL[42].* Centering on gradient leakage, the CPL attack utilizes gradient updates shared during federated learning to reconstruct private data. By iteratively optimizing a noise vector to match observed gradients, this approach enables data reconstruction even without explicit data removal, underscoring risks associated with gradient sharing in federated learning systems.

In the realm of defending against gradient-based attacks in federated learning, existing defenses primarily focus on gradient monitoring or adding computationally expensive cryptographic protocols. To compare the performance of FedANI, we evaluate it against three baseline defense strategies:

*4) Parameter Setting:* Our methods and baseline are implemented in Python 3.9 and utilize the PyTorch library. All experiments are performed on a workstation equipped with one NVIDIA GeForce RTX 4090 GPU. We adopt the following parameter settings for the attacks and defense in the federated unlearning environment: We use the LeNet model for the MNIST and LFW datasets. The LeNet model consists of 3 convolutional layers and 1 fully connected layer. For the CIFAR-10 dataset, we employ the ResNet18 and ConvNet64 models. The ResNet18 model consists of 4 convolutional layers, 3 residual blocks, and 1 fully connected layer. ConvNet64 consists of 7 convolutional layers, 7 batch normalization layers, 7 ReLU activation functions, 2 max-pooling layers, and 1 fully connected layer. In DRAGD/DRAGDP ,we train the models from scratch for a total of 300 iterations, using a fixed learning rate of 0.05 for MNIST/LFW and 0.01 for CIFAR-10, with a batch size of 64. The number of clients per dataset is set to 10. In FedANI, the gradient noise coefficient is set to 1, the variance noise coefficient to 15, and the noise addition (or pruning) ratio $\varphi$ to 0.005. In the non-IID setting, we employ a Dirichlet sampling strategy for data partitioning, where each local client is assigned a portion of the samples for each label based on a Dirichlet distribution with a concentration parameter of 0.1.

### B. Effectiveness of DRAGDP

To assess the effectiveness of the DRAGDP, experiments were conducted under a simulated federated unlearning scenario. For each dataset—MNIST, Cifar-10, and LFW—a subset of images was selected as the complete dataset ($\mathcal{D}_f \cup \mathcal{D}_r$), and gradients computed on this dataset using the FedAvg algorithm served as pre-unlearning gradients. A portion of these images was then extracted as the remaining dataset ($\mathcal{D}_r$), with its corresponding gradients designated as post-unlearning gradients, referred to as the "Part" component. This Part reconstruction was not treated as a standalone algorithm but as a prerequisite step for both DRAGD and DRAGDP, enabling the simulation of gradient-based attacks before and after unlearning. The DLG attack was applied to MNIST and LFW datasets, while the IG attack was used for Cifar-10, serving as baseline methods for comparison.

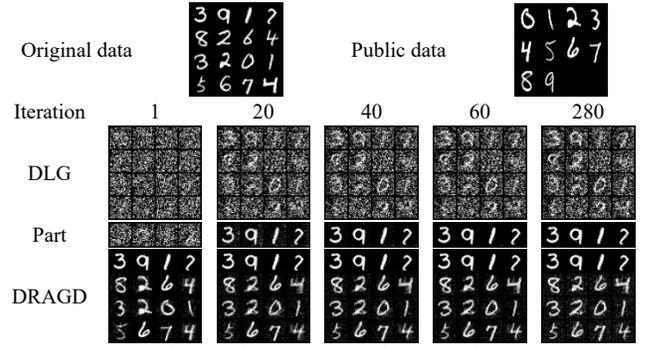

**Fig.2.** Comparison of reconstruction results on MNIST dataset on LeNet model.

Fig.3 illustrates the reconstruction results of the MNIST dataset on the LeNet model. In this experiment, a total of 16 images were selected for reconstruction. After the unlearning process, only 4 images remained in the "Part" subset, corresponding to the retained gradients, while the gradients of the remaining 12 images were forgotten. As the reconstruction progresses, it is observed that the "Part" method yields superior reconstruction quality compared to the DLG approach, primarily due to the reduced number of parameters involved in the optimization. Furthermore, both DRAGD and DRAGDP methods leverage the fixed "Part" images to reconstruct the set of forgotten data. Notably, the images produced by DRAGD and DRAGDP are significantly clearer than those obtained by DLG, demonstrating better overall fidelity. However, occasional reconstruction failures can still be found in DRAGD. In comparison, the DRAGDP method, by utilizing public data as prior knowledge, consistently reconstructs images with sharper details and more accurate features. This highlights the advantage of integrating external prior information into the reconstruction process.

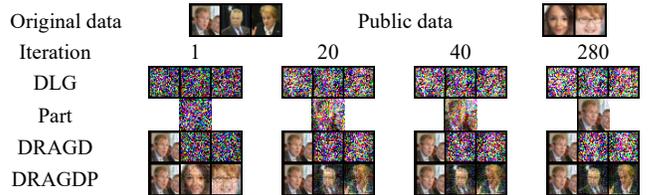

**Fig.3.** Comparison of reconstruction results on LFW dataset on LeNet model.

Next, we conducted a comparable experiment on the LFW dataset using the LeNet model, as shown in Fig.4. Due to the large number of optimization parameters and the limited capacity of the LeNet model—which comprises only three convolutional layers and one fully connected layer—we deliberately selected a smaller attack quantity to manage computational complexity while ensuring the model could carry sufficient gradient information for reconstruction. In this scenario, the attack difficulty of DRAGD was found to be comparable to that of the DLG method, resulting in a failure to reconstruct the forgotten data effectively. Conversely, DRAGDP demonstrated superior attack performance by leveraging prior knowledge of shared facial features, such as the arrangement of facial organs, to initialize the noise data. This approach facilitated more accurate reconstructions, aligning with the consistent findings observed in the MNIST dataset experiments.

TABLE I
COMPARISON OF METRICS FOR IMAGE RECONSTRUCTION ON LENET MODEL

| MNIST dataset | | | | LFW dataset | | | |
|---|---|---|---|---|---|---|---|
| Metrics | MSE↓ | PSNR↑ | SSIM↑ | Metrics | MSE↓ | PSNR↑ | SSIM↑ |
| DLG | 0.539 | 7.21 | 0.218 | DLG | 2.352 | 5.94 | 0.146 |
| Part | 5.18e-5 | 47.34 | 1.000 | Part | 0.002 | 27.18 | 0.963 |
| DRAGD | 0.294 | 18.32 | 0.488 | DRAGD | 1.312 | 13.12 | 0.448 |
| DRAGDP | 0.029 | 24.86 | 0.849 | DRAGDP | 0.011 | 21.40 | 0.879 |

The quantitative results presented in TABLE I clearly demonstrate the superior performance of DRAGDP compared to the baseline methods. Specifically, DRAGDP achieves a substantially lower mean squared error, indicating that its reconstructed images are much closer to the original images at the pixel level. In addition, DRAGDP attains significantly higher peak signal-to-noise ratio and structural similarity index values than both DLG and DRAGD, highlighting its remarkable ability to preserve image structure and visual fidelity.

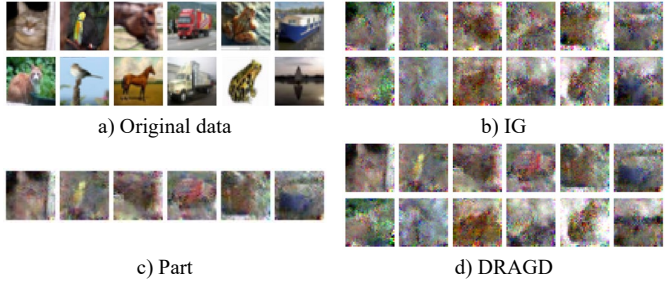

**Fig.4.** Comparison of reconstruction results for Cifar-10 dataset on ConvNet64 model.

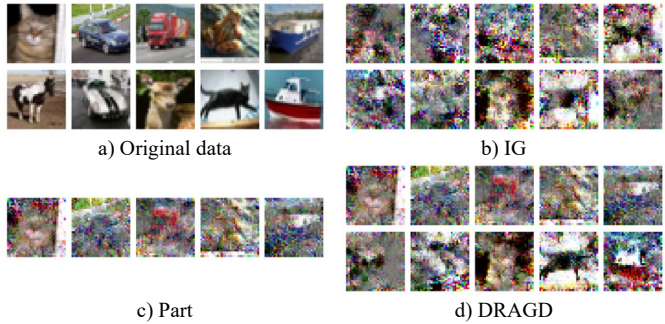

**Fig.5.** Comparison of reconstruction results for Cifar-10 dataset on ResNet18 model.

TABLE II
COMPARISON OF METRICS FOR IMAGE RECONSTRUCTION ON CIFAR-10 DATASET

| ConvNet64 | | | | ResNet18 | | | |
|---|---|---|---|---|---|---|---|
| Metrics | MSE↓ | PSNR↑ | SSIM↑ | Metrics | MSE↓ | PSNR↑ | SSIM↑ |
| IG | 0.668 | 14.18 | 0.105 | IG | 1.528 | 10.36 | 0.068 |
| Part | 0.260 | 18.12 | 0.305 | Part | 0.633 | 14.09 | 0.265 |
| DRAGD | 0.358 | 16.86 | 0.265 | DRAGD | 0.791 | 13.27 | 0.301 |

The attack results on the Cifar-10 dataset using ConvNet64 and ResNet18 models are illustrated in Fig.5 and Fig.6, with the corresponding quantitative image similarity metrics summarized in TABLE II. Compared to the previous experiments on the simpler LeNet model, attacks targeting the gradients of these more complex models are able to reconstruct images on a larger scale and with greater structural complexity. Both the visual inspection of the reconstructed images and the numerical evaluation metrics consistently demonstrate that gradient-based attacks on advanced network architectures yield more detailed and intricate reconstructions, highlighting the increased vulnerability of complex models to such privacy attacks.

*C. Ablation Study*

In this section, we conduct ablation studies to systematically investigate the impact of prior knowledge integration within the DRAGDP method and to examine the role of fixed Part images in both the DRAGD and DRAGDP algorithms. Specifically, these experiments aim to clarify how the inclusion of external prior information influences the reconstruction quality, as well as to elucidate the significance of anchoring certain images after the unlearning process. Through this analysis, we seek to provide deeper insights into the contributing factors that enhance the effectiveness and robustness of gradient-based data reconstruction attacks.

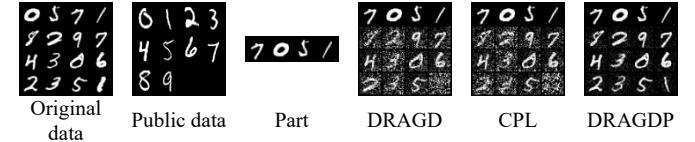

Original data | Public data | Part | DRAGD | CPL | DRAGDP

**Fig.6.** Comparison of reconstruction results with different prior knowledge on MNIST dataset.

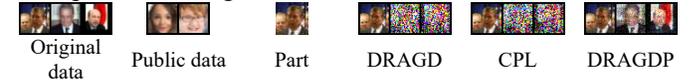

Original data | Public data | Part | DRAGD | CPL | DRAGDP

**Fig.7.** Comparison of reconstruction results with different prior knowledge on LFW dataset.

TABLE III
COMPARISON OF METRICS FOR IMAGE RECONSTRUCTION WITH DIFFERENT PRIOR KNOWLEDGE

| MNIST | | | | LFW | | | |
|---|---|---|---|---|---|---|---|
| Metrics | MSE↓ | PSNR↑ | SSIM↑ | Metrics | MSE↓ | PSNR↑ | SSIM↑ |
| Part | 1.72e-4 | 48.32 | 0.998 | Part | 0.004 | 23.85 | 0.953 |
| DRAGD | 0.209 | 19.87 | 0.631 | DRAGD | 1.405 | 12.08 | 0.510 |
| CPL | 0.199 | 19.91 | 0.628 | CPL | 0.887 | 12.30 | 0.521 |
| DRAGDP | 0.020 | 26.50 | 0.899 | DRAGDP | 0.022 | 18.45 | 0.867 |

To begin, we evaluate the effectiveness of incorporating prior knowledge by comparing DRAGDP with the CPL algorith. In the CPL method, one quarter of the original image is randomly initialized with noise and then replicated across the remaining three sections as the initial input for reconstruction. As illustrated in Fig.7 and Fig.8, the introduction of prior knowledge in DRAGDP leads to a substantial improvement in reconstruction quality over both DRAGD and CPL. The quantitative image similarity metrics, summarized in TABLE III, further support these observations. While CPL demonstrates enhanced attack performance compared to DRAGD, there remains a significant gap between CPL and DRAGDP, underscoring the critical role of effective prior knowledge in achieving high-fidelity data reconstruction.

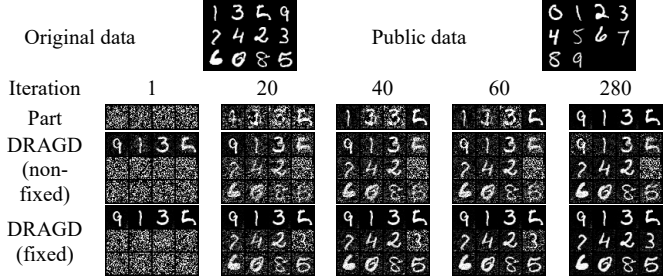

**Fig.8.** Comparison between non-fixed and fixed DRAGD on MNIST dataset.

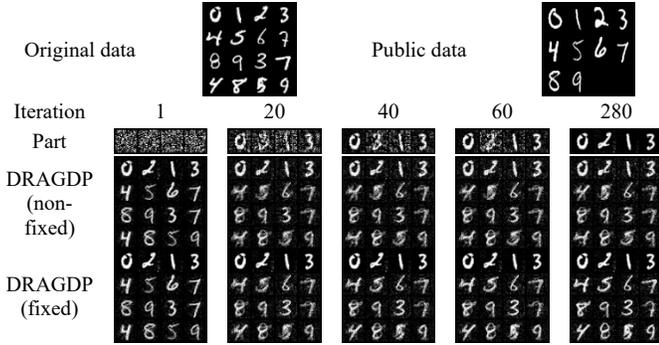

**Fig.9.** Comparison between non-fixed and fixed DRAGDP on MNIST dataset.

TABLE IV
COMPARISON OF METRICS FOR NON-FIXED AND FIXED DRAGD/DRAGDP ON MNIST DATASET

| | DRAGD | | | | DRAGDP | | |
|---|---|---|---|---|---|---|---|
| Metrics | MSE↓ | PSNR↑ | SSIM↑ | Metrics | MSE↓ | PSNR↑ | SSIM↑ |
| non-fixed | 0.144 | 14.42 | 0.713 | non-fixed | 0.036 | 16.24 | 0.812 |
| fixed | 0.006 | 29.62 | 0.975 | fixed | 0.016 | 23.39 | 0.915 |

Next, we investigate the impact of fixing the parameters of $\mathcal{N}_r$, that is, anchoring the Part images during the reconstruction process, on the performance of DRAGD and DRAGDP. To this end, we relax the constraint of fixed Part images and allow all data, including those in the Part set, to be updated during training. Experiments are conducted on both the MNIST datasets, as illustrated in Fig.9 and Fig.10. The results clearly indicate that, compared to scenarios with fixed Part images, the reconstruction quality of both DRAGD and DRAGDP degrades significantly when the Part data are not fixed. In particular, the DRAGD method exhibits a trend in which the discrepancy between the reconstructed Part images and the ground truth first increases and then gradually decreases. This phenomenon can be attributed to the initial adaptation to noise, which reduces pixel quality, followed by subsequent optimization that, however, fails to fully restore the original reconstruction performance. These observations suggest that removing the fixed constraint leads to an over-reliance on pre-unlearning gradients, causing the optimization objective to converge to suboptimal local minima. Quantitative analysis, as summarized in TABLE IV, further confirms the importance of fixing Part data to ensure effective and stable attack performance.

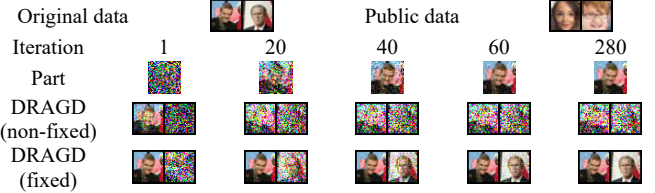

**Fig.10.** Comparison between non-fixed and fixed DRAGD on LFW dataset.

## VII. CONCLUSION

In this paper, we have highlighted the privacy vulnerabilities within FU by introducing novel gradient-based attack methods that exploit discrepancies in gradients before and after data removal. We propose DRAGD, a data reconstruction attack that systematically targets these gradient differences to reconstruct forgotten data, and its enhanced version, DRAGDP, which leverages publicly available data to improve the attack's effectiveness, particularly in tasks with strong prior structures, such as facial recognition. To address these privacy risks, we introduce FedANI, an adaptive and lightweight defense mechanism that selectively injects noise into sensitive gradient components, preventing data reconstruction without compromising the federated model's utility. Extensive evaluations on benchmark datasets such as MNIST, CIFAR-10, and LFW demonstrate that DRAGD and DRAGDP significantly outperform previous attacks in terms of data reconstruction accuracy, while FedANI provides robust defense capabilities, maintaining the balance between privacy protection and model performance. This work not only exposes critical security risks in federated unlearning but also offers a practical solution that enhances privacy-preserving federated learning systems.